\documentclass[10pt,twocolumn,letterpaper]{article}

\usepackage[]{cvpr}      

\usepackage{algorithmic}
\usepackage{algorithm}
\usepackage{arydshln}
\usepackage{multirow}
\usepackage{makecell}
\usepackage{graphicx}
\definecolor{custompink}{RGB}{255,105,180} 
\definecolor{cvprblue}{rgb}{0.21,0.49,0.74}
\usepackage[pagebackref,breaklinks,colorlinks,allcolors=cvprblue]{hyperref}

\def\paperID{xxx} 
\def\confName{CVPR}
\def\confYear{2026}

\author{
Chenyang Wu\textsuperscript{1,3}\quad Lina Lei\textsuperscript{1,3}\quad Fan Li\textsuperscript{2} \quad Chunle Guo\textsuperscript{1,3 $\dag$} \quad Dehong Kong\textsuperscript{2} \quad Xinran Qin\textsuperscript{2} \\ \quad Zhixin Wang\textsuperscript{2} \quad Mingming Cheng\textsuperscript{1,3} \quad Chongyi Li\textsuperscript{1,3}\\
\textsuperscript{1}VCIP, CS, Nankai University\quad
\textsuperscript{2}Huawei Noah’s Ark Lab\quad
\textsuperscript{3}NKIARI, Shenzhen Futian
\\
\{chenyangwu, leilina\}@mail.nankai.edu.cn, \\
\{lifan61, kongdehong3, qinxinran, wangzhixin6\}@huawei.com\\
    \{guochunle, cmm, lichongyi\}@nankai.edu.cn \\
}

\begin{document}

\title{YOSE: You Only Select Essential Tokens for Efficient \\ DiT-based Video Object Removal}

\twocolumn[{
\renewcommand\twocolumn[1][]{#1}
\maketitle
\begin{center}
    \captionsetup{type=figure}
    \includegraphics[width=0.95\textwidth]{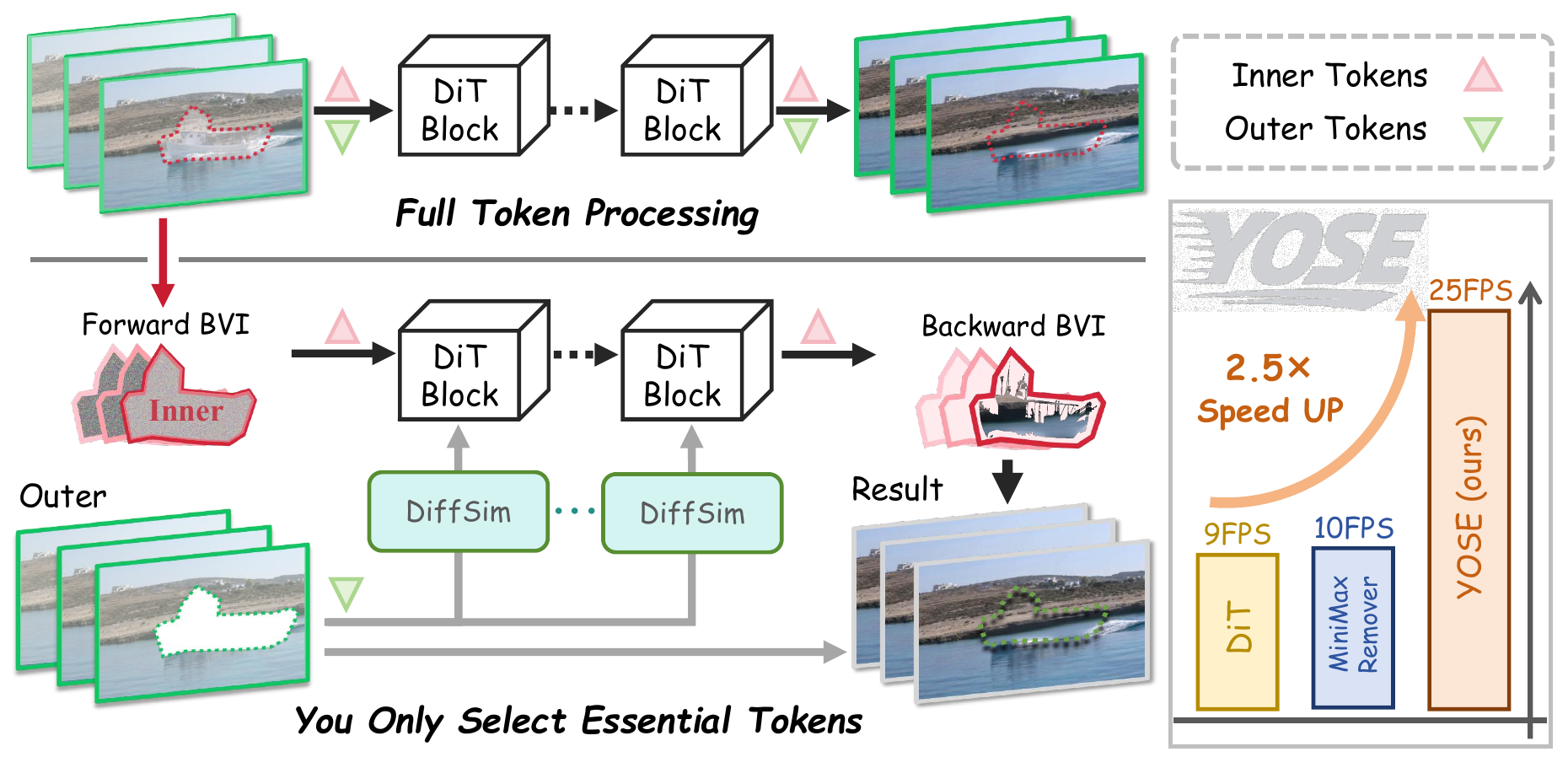}
    \captionof{figure}{\textbf{Overview of the proposed YOSE. }
    \textit{`Forward / Backward BVI'} denotes the Forward / Backward Batch Variable-length Indexing (BVI). 
    `DiffSim' means the Diffusion Process Simulator module. 
    Given a masked video, YOSE employs BVI to selectively process only essential tokens within masked areas, avoiding redundant computation (Sec.~\ref {sec:bvi}).
    Then, DiffSim simulates the diffusion process influence of unmasked regions to preserve semantic consistency (Sec.~\ref{sec:prosim}).
    Finally, by aligning mean–variance distributions in overlapping regions, YOSE seamlessly fuses restored and original areas, effectively eliminating boundary artifacts while maintaining high visual quality (Sec.~\ref{subsec:fusion}).
    The FPS results are tested at 480p resolution.}
    \label{fig:teaser}
\end{center}
}]

\let\thefootnote\relax\footnotetext{
$\dag$ Corresponding Author. 
}

\begin{abstract}
Recent advances in Diffusion Transformer (DiT)-based video generation technologies have shown impressive results for video object removal. 
However, these methods still suffer from substantial inference latency.
For instance, although MiniMax Remover achieves state-of-the-art visual quality, it operates at only around 10FPS, primarily due to dense computations over the entire spatiotemporal token space, even when only a small masked region actually requires processing.
In this paper, we present \textbf{YOSE}, \textbf{Y}ou \textbf{O}nly \textbf{S}elect \textbf{E}ssential Tokens, an efficient fine-tuning framework.
YOSE introduces two key components: Batch Variable-length Indexing (BVI) and Diffusion Process Simulator (DiffSim) Module. 
BVI is a differentiable dynamic indexing operator that adaptively selects essential tokens based on mask information, enabling variable-length token processing across samples.
DiffSim provides a diffusion process approximation mechanism for unmasked tokens, which simulates the influence of unmasked regions within DiT self-attention to maintain semantic consistency for masked tokens.
With these designs, YOSE achieves mask-aware acceleration, where the inference time scales approximately linearly with the masked regions, in contrast to full-token diffusion methods whose computation remains constant regardless of the mask size.
Extensive experiments demonstrate that YOSE achieves up to 2.5$\times$ speedup in 70\% of cases while maintaining visual quality comparable to the baseline.
Code is available at: \textcolor{custompink}{\texttt{https://github.com/Wucy0519/YOSE-CVPR26}}.

\end{abstract}

\section{Introduction}
\label{sec:intro}

Recent advances in DiT have significantly boosted the performance of visual generation tasks~\cite{esser2024scaling, wan2025wan, kong2025dual, zhong2025outdreamer}.
DiTs leverage the global modeling capability of transformers, enabling stronger long-range dependency learning and superior scalability across spatial and temporal dimensions~\cite{wan2025wan}.
Recent large-scale video diffusion systems, such as VideoDiT~\cite{fengvideodit}, Wan 2.1~\cite{wan2025wan}, and VACE~\cite{jiang2025vace}, demonstrate that DiT architectures can produce high-quality, temporally consistent videos across diverse conditions, including text prompts, reference frames, and masks.

Video object removal (VOR) aims to remove undesired objects from videos while maintaining spatial realism and temporal consistency.
Benefiting from the development of the DiT-based diffusion models, VOR has achieved remarkable progress in recent years~\cite{jiang2025vace, zi2025minimax, miao2025rose}.
However, these methods remain inherently inefficient during inference. 
Among them, MiniMax Remover~\cite{zi2025minimax}, as a state-of-the-art method, leverages flow-matching diffusion for coherent video reconstruction.
Despite its strong generative ability, MiniMax Remover inherits the DiT architecture’s full-token computation paradigm, performing dense denoising and attention operations over the entire spatiotemporal token space, even though only masked regions require processing.
This results in significant computational redundancy and high inference latency, as the complexity remains constant regardless of the mask size.

In practice, VOR is a localized generation task, i.e., only the masked regions need to be processed, while the unmasked regions remain unchanged. 
Around 70\% of VOR cases have mask ratios below 20\% (statistics on YouTuBe-VOS~\cite{DBLP:journals/corr/abs-1809-03327} and DAVIS~\cite{DBLP:conf/cvpr/PerazziPMGGS16} datasets, which are widely used as benchmarks, e.g. VFI~\cite{wu2025vtinker}), suggesting that substantial acceleration is achievable if computation scales with the masked regions.
However, existing DiT-based VOR models, represented by Minimax Remover, perform denoising and attention over all tokens, leading to redundant computation and high inference latency.
This inefficiency becomes even more pronounced when the masked regions are small, as the computational cost remains fixed regardless of the actual mask size.

To address this problem, we propose \textbf{YOSE} — \textbf{Y}ou \textbf{O}nly \textbf{S}elect \textbf{E}ssential Tokens, a simple yet effective fine-tuning framework.
Rather than re-designing the model architecture, YOSE is a lightweight and plug-in framework that optimizes MiniMax Remover for mask-aware computation.
The core of YOSE is to process only the essential tokens corresponding to the masked regions, while preserving the semantic influence of the unmasked region through contextual simulation.

\begin{figure}[t]
  \centering
   \includegraphics[width=1\linewidth]{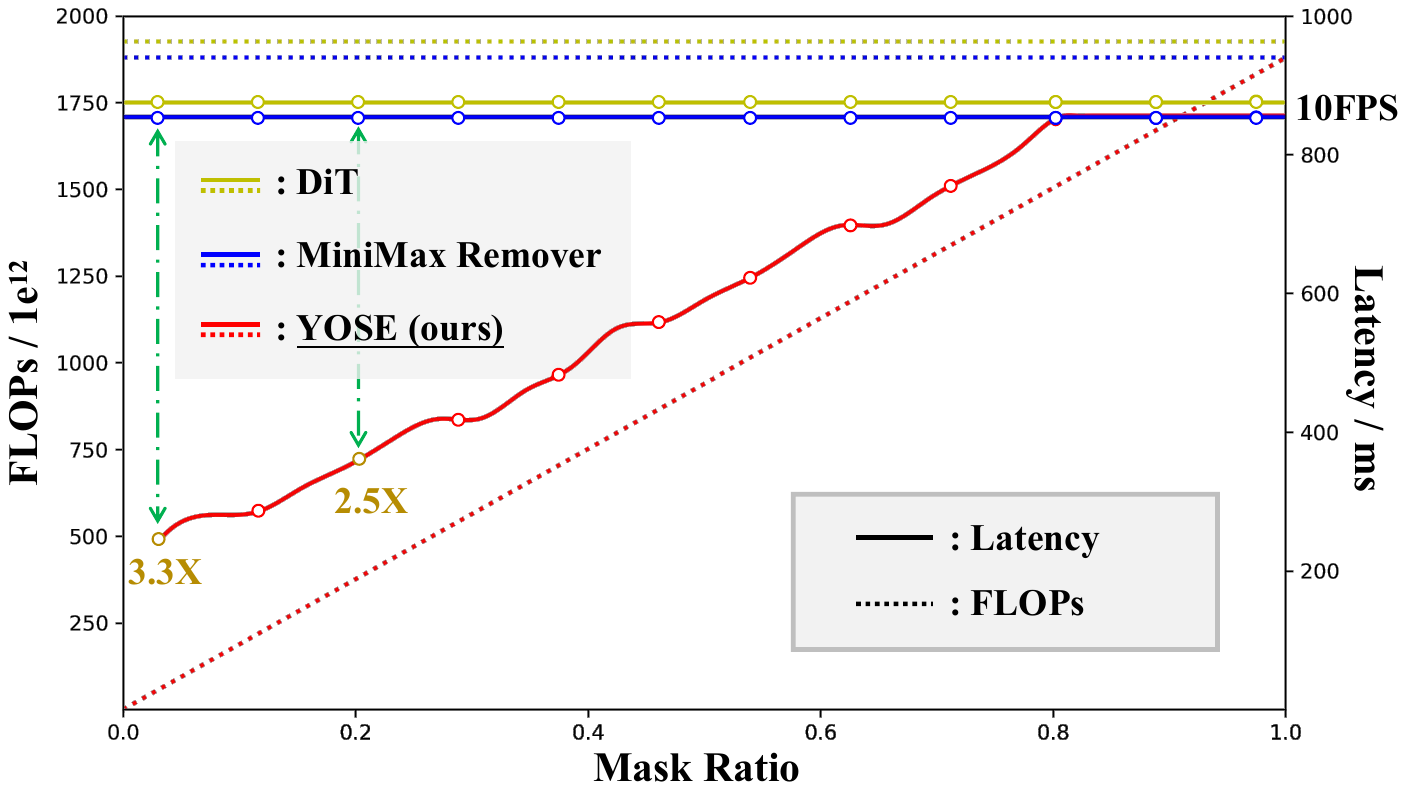}
   \caption{Comparison among DiT, MiniMax Remover, and YOSE (use Minimax Remover as base model). 
   MiniMax Remover adopted the design of DiT without cross attention.
   We calculate the FLOPs and Latency of the multiple DiT Blocks in the case of processing an 81-frame video at 480P resolutions (more detail in Eq.~\eqref{eq:flops2}). 
   As shown above, the FLOPs (theoretical curve) and Latency of YOSE grow linearly with the mask ratio, enabling faster inference for smaller masked regions, about 25FPS for 20\% mask ratio, without affecting processing outcomes (see Tab.~\ref{tab:maintab}).}
   \label{fig:fig2}
\end{figure}

Specifically, as shown in Fig.~\ref{fig:teaser}, YOSE introduces two complementary components:
Batch Variable-length Indexing (BVI) and Diffusion Process Simulator (DiffSim) Module.
BVI is a differentiable dynamic indexing operator that adaptively selects tokens based on mask information, allowing variable-length token processing across samples while maintaining gradient flow.
DiffSim is a lightweight diffusion process modeling module for unmasked (outer) regions, which simulates their influence within each DiT block’s self-attention, ensuring semantic consistency across mask boundaries.
Together, these designs transform MiniMax Remover into a mask-aware, token-efficient diffusion framework with linear computational complexity, where inference time scales proportionally with the mask region. 
As shown in Fig.~\ref{fig:fig2}, incorporating YOSE into MiniMax Remover obtains a significant reduction in FLOPs and latency, achieving up to 2.5$\times$ speedup for videos at 480P resolutions, while maintaining comparable visual quality.
Furthermore, we also applied YOSE to VACE~\cite{jiang2025vace}, a ControlNet-like DiT-based video editing model, demonstrating its generalization.

In summary, our main contributions are as follows:
\begin{itemize}
\item We propose YOSE, a mask-aware fine-tuning framework, which accelerates DiT-based video object removal, e.g., MiniMax Remover, without sacrificing visual quality.

\item We introduce the Batch Variable-length Indexing (BVI), which dynamically and differentiably selects only the masked-region tokens based on the mask information, enabling variable-length and efficient token processing across samples.

\item We present the Diffusion Process Simulator Module, which simulates the influence of the unmasked regions' diffusion process within each DiT block, ensuring semantic and temporal consistency without full-token computation.

\end{itemize}

\begin{figure*}[t]
    \centering
    \includegraphics[width=0.95\textwidth]{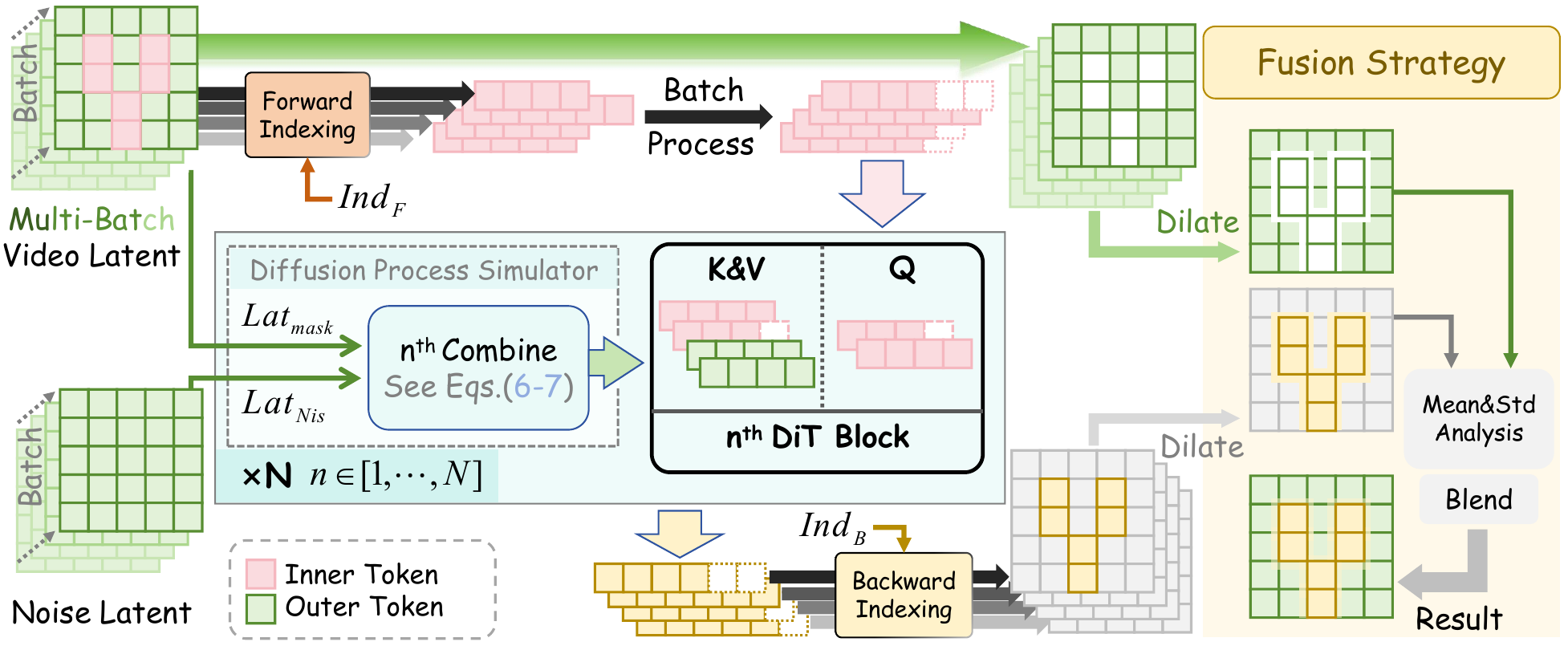} 
    \caption{Framework of the proposed YOSE. 
    ${Lat_{mask}}$ and ${Lat_{Nis}}$ denote the latent features of masked video and noise. 
    Given a masked video, YOSE first performs Batch Variable-length Indexing (BVI) to dynamically select essential tokens corresponding to the masked regions (Inner), reducing redundant computation. 
    Then, to simulate the influence of the outer tokens' diffusion process and maintain semantic consistency with the inner, the Diffusion Process Simulator (DiffSim) introduces the influence of outer tokens through specific simulation during each DiT block stage. 
    Finally, by creating an overlapping region between the inner and outer areas, aligning their mean and variance (std) distributions, and blending them with a weighted mask, YOSE effectively smooths the transition and eliminates visible boundary artifacts.
    By employing BVI and DiffSim, YOSE achieves competitive visual quality and inference speed.}%
    \label{fig:framework}
\end{figure*}

\section{Related Work}
\label{sec:relatedwork}

Recent progress in DiT has significantly advanced generative modeling across image~\cite{duan2025dit4sr, feng2025dit4edit} and video~\cite{wan2025wan, jiang2025vace, zi2025minimax, bian2025videopainter} domains.
Unlike UNet-based diffusion architectures, which rely on convolutional inductive biases, DiTs reformulate the diffusion denoising process as a sequence-to-sequence transformation on latent tokens~\cite{wan2025wan}.
This design unifies diffusion and transformer paradigms, offering superior scalability and representation capacity for high-resolution generation~\cite{esser2024scaling}.
Following DiT’s success in images, several studies~\cite{yang2024cogvideox, wan2025wan, fengvideodit} have extended it to video generation.
Models such as VideoDiT~\cite{fengvideodit} and Wan 2.1~\cite{wan2025wan} integrate spatial–temporal attention to model motion dynamics and visual fidelity jointly.

While prior works have extensively explored image~\cite{lu2025grig, qin2025unified, qin2026disentangle, qin2025robust, li2024magiceraser, meng2025ultraled} \& video~\cite{zhou2023propainter} editing, recent DiT-based architectures have emerged as powerful tools for the more complex task~\cite{qin2025camedit, duan2025dit4sr}.
Building upon Wan 2.1, ROSE~\cite{miao2025rose} and Minimax Remover~\cite{zi2025minimax} leverage diffusion transformers for video object removal, demonstrating that DiT-based models can learn both spatial restoration and temporal coherence within a unified generative framework.
ROSE~\cite{miao2025rose} introduces a DiT-based framework for video object removal that explicitly models object-induced environmental effects, such as shadows, reflections, and translucency, through synthetic paired data generated via 3D rendering, enabling comprehensive side-effect-aware object erasure and improved generalization to real-world videos.
Minimax Remover~\cite{zi2025minimax} is a two-stage DiT-based VOR framework, which employs a minimax distillation strategy to improve robustness and visual quality, enabling state-of-the-art removal performance with as few as 6 sampling steps and without relying on classifier-free guidance.
Since ROSE does not rely on a fixed mask, applying our method directly to it presents challenges.
Therefore, we mainly use Minimax Remover~\cite{zi2025minimax} as the base model to demonstrate the effectiveness of YOSE.
We also validate YOSE's generalizability on other base models, e.g., VACE.

\section{Methodology}
\label{sec:method}

As shown in Fig.~\ref{fig:framework}, given a multi-batch masked video ${Video_{mask}}$, to dynamically select the essential tokens within the masked regions (inner), YOSE first performs Batch Variable-length Indexing (BVI, as described in Alg.~\ref{alg:batchindex}) to obtain the forward BVI index ${Ind_{F}}$ and backward BVI index ${Ind_{B}}$. 
Then, via the forward BVI index ${Ind_{F}}$, we obtain a token sequence ${St_{in}}$ from the masked region, which will be processed in DiT. 
After that, in each DiT block, to simulate the diffusion process influence of the outer region and ensure semantic consistency with the inner, the Diffusion Process Simulator (DiffSim) module (pseudocode is detailed in the supplementary materials) introduces three learnable parameters, i.e., combining parameter ${\mathcal{G}}$, scaling parameter ${\mathcal{S}}$, and bias parameter ${\mathcal{B}_{ias}}$. 
Finally, through the backward BVI index ${Ind_{B}}$, the results ${St_{in}^{out}}$ are applied to the areas requiring removal.

\begin{algorithm}[t]
    \caption{Batch Variable-length Indexing}
    \label{alg:batchindex}
    \renewcommand{\algorithmicrequire}{\textbf{Input:}}
    \renewcommand{\algorithmicensure}{\textbf{Output:}}
    
    \begin{algorithmic}[1]
        \REQUIRE $mask$
        \ENSURE ${Ind_{F}}$, ${Ind_{B}}$
        
        \STATE ${\mathcal{L}}$: Number of Tokens in each batch
        \STATE ${\mathcal{B}}$: Number of Batches
        \STATE ${Ind_{F}}$, ${Ind_{B}}$: Forward/Backward Batch Variable- length Index
        \STATE ${Length()}$: Function for Counting the Number of Elements. i.e. ${\mathcal{L}=Length(mask[0])}$
        \STATE ${Linspace()}$: Function `Torch.Linspace' in Torch
        \STATE ${Sample=Linspace(1/2\mathcal{L}-1, 1-1/2\mathcal{L}, \mathcal{L})}$
        \STATE ${Ind_F}$, ${Ind_B}$ = ${[\quad]}$ , ${[\quad]}$
        \STATE ${Len_F}$ = ${[\quad]}$
        \STATE /* For Forward Index ${Ind_{F}}$ */
        \FOR{each $i \in [0, \mathcal{B}-1]$}
            \STATE ${This\_Ind = Sample[mask[i]]}$
            \STATE ${Ind_F = Ind_F \cup [This\_Ind]}$
            \STATE ${Len_F = Len_F \cup [length(This\_Ind)]}$
        \ENDFOR

        \STATE ${\mathcal{L}_{max} = Max(Len_F)}$
        \STATE ${\Delta_{max}=1/2\mathcal{L}_{max}}$

        \STATE /* For Backward Index ${Ind_{B}}$ */
        \FOR{each $i \in [0, \mathcal{B}-1]$}
            \STATE ${Ind\_Short = Linspace(\Delta_{max}-1,\Delta_{max}-1+}$
            \STATE ${\qquad  (Len_F[i]-1)\times(2-2\Delta_{max})/(\mathcal{L}_{max}-1)}$
            \STATE ${\qquad ,Len_F[i])}$
            \STATE ${tmp_{B}=Sample}$
            \STATE ${tmp_{B}[mask[i]]=Ind\_Short}$
            \IF {$Len_F[i]\quad!=\quad \mathcal{L}_{max}$}
                \STATE ${Padd=[1, ..., 1]}$
                \STATE ${Ind_F[i] = Ind_F[i]\cup Padd[:\mathcal{L}_{max}-Len_F[i]]}$
            \ENDIF
            \STATE ${Ind_B=Ind_B\cup [tmp_B]}$
        \ENDFOR
        
        \RETURN ${Ind_{F}}$, ${Ind_{B}}$
    \end{algorithmic}
\end{algorithm}

\subsection{Batch Variable-length Indexing}
\label{sec:bvi}

To retrieve needed tokens based on the mask, a common approach is as follows.
First, the selected tokens ${St_{in}}$ are mapped by performing an address retrieval operation,
\begin{equation}
    St_{in} = Video_{mask}[mask].
\label{eq:eq0}
\end{equation}
After processing by the multi-DiT model, ${St_{in}^{out}}$ is obtained.
Based on the partial result ${St_{in}^{out}}$, to achieve a complete output, the results ${St_{in}^{out}}$ are mapped back to their original locations based on the mask, which is represented by
\begin{equation}
 Video_{mask}[mask] = St_{in}^{out}.
\end{equation}
However, this approach faces two core challenges: gradient calculations during forward/backward propagation and batch training/inference due to the different numbers of masked tokens in the batch, which limits the efficiency of the model.

To address these challenges, we propose the Batch Variable-length Indexing (BVI) algorithm, which enables differentiable and efficient token selection according to the given mask while supporting variable-length batches across samples.
Specifically, BVI constructs two types of indices: the forward index ${Ind_F}$, which selects the set of masked tokens to be processed, and the backward index ${Ind_B}$, which restores the processed results back to their original spatial-temporal positions.
This design ensures that both the forward and backward propagation can be conducted seamlessly within the diffusion process.

The core of BVI lies in transforming discrete mask-based sampling into a differentiable coordinate-mapping operation.
Instead of performing hard indexing, such as ${Tensor[mask]}$ in the Eq.~\eqref{eq:eq0}, which blocks gradient flow, BVI employs the grid sample function (referred to as `GSample()' below) to realize continuous interpolation-based selection. 
The selection of masked tokens can be represented as a differentiable sampling function, enabling end-to-end optimization.
Formally, given a normalized coordinate grid ${Ind_F}$ , we define the token extraction as:
\begin{equation}
St_{in} = \mathrm{GSample}(Video_{mask}, Ind_F),
\label{eq:eq1}
\end{equation}
where gradients can propagate through both the video feature and the sampling coordinates.
After diffusion-based processing, the tokens are restored via the backward index ${Ind_B}$ in the reverse direction:
\begin{equation}
Video_{out} = \mathrm{GSample}(St_{in}^{out}, Ind_B),
\label{eq:eq2}
\end{equation}
thus recovering the full video with updated masked regions.

A key advantage of BVI is its ability to handle varying mask sizes within a batch.
Unlike standard diffusion models, where the token length is fixed, BVI dynamically computes the token counts per sample and pads them to the maximum token length within the batch.
This enables efficient batch parallelization while avoiding unnecessary computation on unmasked regions.
Consequently, the overall computational complexity becomes proportional to the number of masked tokens rather than the total number of video tokens, enabling YOSE to achieve linear complexity with respect to mask size.
The pseudocode of BVI can be found in Alg.~\ref{alg:batchindex}.

\subsection{Diffusion Process Simulator Module}
\label{sec:prosim}

Although BVI enables efficient and differentiable selection of the masked-region tokens, directly processing only these selected tokens may result in the inability to model contextual dependencies between inner and outer regions — a critical limitation for video object removal tasks, where semantic continuity and motion coherence across the boundary are essential.
To address this, we introduce the Diffusion Process Simulator (DiffSim) module, which explicitly simulates the diffusion process influence of the unmasked regions during DiT-based generation, while avoiding redundant computation on the full token space.

First, the DiffSim embeds the input mask according to 3D-VAE's strides (will be described in the supplementary materials for details).
Based on the embedded mask ${mask}$, we construct two complementary index sets using BVI: the inner-region indices $(Ind_{F\_in}, Ind_{B\_in})$ and the outer-region indices $(Ind_{F\_out}, Ind_{B\_out})$.
For the processing of outer tokens in full-token DiT, the model's input is the noised video latent, and its prediction is noise, both of which are known when inference. 
We assume that the token's state after some DiT processing is simultaneously related to both the noise latent ${Lat_{Nis}}$ and the masked video latent ${Lat_{mask}}$.
Based on the assumption above, DiffSim introduces a loss-inspired (here is flow-matching loss) residual latent representation ${Res_{Nis}}$:
\begin{equation}
{Res_{Nis}=Lat_{Nis}-Lat_{mask}}.
\label{eq:maineq}
\end{equation}
where ${Res_{Nis}}$ is specifically designed to simulate the intermediate state of external tokens within each DiT processing.
Intuitively, ${Res_{Nis}}$ captures how the noise evolves toward the target latent state under the learned flow field.
Furthermore, we suggest that combining ${Res_{Nis}}$ and ${Lat_{mask}}$ can simulate the intermediate states of these unmasked tokens.
After combining ${Res_{Nis}}$ and ${Lat_{mask}}$, DiffSim can generate proxy key\&value (KV) features that approximate the diffusion process influence of the outer (unmasked) regions without explicitly processing them through DiT.
This design ensures that, within each DiT block, the inner tokens and the simulated outer tokens reside in a unified latent domain, allowing the inner tokens to readily attend to the outer context and maintain consistency across the masked boundary.

In each DiT block, DiffSim introduces three learnable parameter groups: the combining parameter ${\mathcal{G}}$, scaling parameter ${\mathcal{S}}$, and bias parameter ${\mathcal{B}{ias}}$.
These parameters dynamically adjust the degree of contextual blending and modulation of the simulated unmasked regions' features.
Specifically, the combining parameter ${\mathcal{G}}$ determines the interpolation ratio between the reconstructed latent (${Lat_{mask}}$) and its residual component (${Res_{Nis}}$), forming a contextual KV representation:
\begin{equation}
KV = \mathcal{G}[i] \cdot Lat_{mask} + (1 - \mathcal{G}[i]) \cdot Res_{Nis}.
\label{eq:kv}
\end{equation}
Subsequently, the scaling parameter ${\mathcal{S}}$ and bias parameter ${\mathcal{B}{ias}}$ modulate the distribution of ${KV}$ before the attention operation, allowing the model to learn adaptive simulation across each DiT steps:
\begin{equation}
KV = (1 + \mathcal{S}[i]) \cdot KV + \mathcal{B}{ias}[i].
\label{eq:mod}
\end{equation}
Through this process, the simulated keys and values mimic the response of unmasked regions, ensuring that attention computations inside DiT remain aware of global semantics without incurring additional token-level computation.
Finally, after all DiT processing, through the backward BVI mapping by ${Ind_{B\_in}}$, the complete result ${out}$:
\begin{equation}
out = mask*GSample(St_{in}^{out}, Ind_{B\_in}).
\end{equation}

\begin{table*}[t]
  \centering
  \scalebox{0.7}[0.78]
  {
\begin{tabular}{ccclllllll}
\toprule
\multirow{2}[4]{*}{Dataset} & \multirow{2}[4]{*}{Method} & \multirow{2}[4]{*}{Venue} & \multicolumn{4}{c}{VBench~\cite{huang2024vbench} Evaluation} & \multicolumn{3}{c}{Background} \\
\cmidrule(r){4-7} \cmidrule(r){8-10}     &       &       & Mot.Smo.${\uparrow}$ & Dyn.Deg.${\uparrow}$ & Aes.Qua.${\uparrow}$ & Ima.Qua.${\uparrow}$ & PSNR${\uparrow}$ & SSIM${\uparrow}$ & LPIPS${\downarrow}$  \\
\midrule
\multirow{7}[6]{*}{\rotatebox{90}{\qquad YouTube-VOS~\cite{DBLP:journals/corr/abs-1809-03327}}} & Propainter~\cite{zhou2023propainter} & ICCV23& 0.9641  & 0.6667  & 0.3799  & 0.5519  & 41.1438  & 0.9905  & 0.0077   \\
      & DiffuEraser~\cite{li2025diffueraser} & $-$   & 0.9664  & 0.6889  & 0.3717  & 0.5528  & 34.3382  & 0.9634  & 0.0306  \\
      & ROSE~\cite{miao2025rose} & NeurIPS25& 0.9671  & 0.7778  & 0.3916  & 0.5449  & 26.7564  & 0.8708  & 0.0859    \\
\cdashline{2-10}[1pt/1pt]      & VACE~\cite{jiang2025vace} & ICCV25& 0.9714  & 0.6667  & 0.4127  & 0.5517  & 23.7225  & 0.8322  & 0.1322  \\
      & YOSE (VACE) & Ours & 0.9666 \textcolor{blue}{${\downarrow <1e^{-2}}$}  & 0.6705 \textcolor{blue}{${\uparrow <1e^{-2}}$} & 0.3898 \textcolor{red}{${\downarrow0.02}$}  & 0.5277 \textcolor{red}{${\downarrow 0.02}$}  & 29.1917 \textcolor{green!70!black}{${\uparrow5.47}$} & 0.8994 \textcolor{green!70!black}{${\uparrow0.07}$} & 0.0746 \textcolor{green!70!black}{${\uparrow0.06}$}  \\
\cdashline{2-10}[1pt/1pt]      & MiniMax~\cite{zi2025minimax} & NeurIPS25& 0.9668  & 0.6667  & 0.3920  & 0.5361  & 30.3296  & 0.9116  & 0.0615  \\
      & YOSE (Minimax) & Ours & 0.9676 \textcolor{blue}{${\uparrow <1e^{-2}}$} & 0.6778 \textcolor{green!70!black}{${\uparrow0.01}$} & 0.3927 \textcolor{blue}{${\uparrow <1e^{-2}}$} & 0.5298 \textcolor{red}{${\downarrow0.01}$}  & 31.0117 \textcolor{green!70!black}{${\uparrow0.68}$} & 0.9120 \textcolor{blue}{${\uparrow <1e^{-2}}$} & 0.0642 \textcolor{blue}{${\downarrow <1e^{-2}}$}   \\
\midrule
\multirow{7}[6]{*}{\rotatebox{90}{   DAVIS~\cite{DBLP:conf/cvpr/PerazziPMGGS16}}} & Propainter~\cite{zhou2023propainter} & ICCV23& 0.9748  & 0.5889  & 0.4334  & 0.6026  & 39.5823  & 0.9888  & 0.0089  \\
      & DiffuEraser~\cite{li2025diffueraser} & $-$   & 0.9749  & 0.5889  & 0.4365  & 0.6062  & 33.6515  & 0.9549  & 0.0367  \\
      & ROSE~\cite{miao2025rose} & NeurIPS25& 0.9749  & 0.7111  & 0.4485  & 0.6083  & 26.5143  & 0.8286  & 0.1028 \\
\cdashline{2-10}[1pt/1pt]      & VACE~\cite{jiang2025vace} & ICCV25& 0.9769  & 0.5889  & 0.4481  & 0.6019  & 25.0868  & 0.8221  & 0.1228 \\
      & YOSE (VACE) & Ours & 0.9746 \textcolor{blue}{${\downarrow <1e^{-2}}$}  & 0.5778 \textcolor{red}{${\downarrow0.01}$}  & 0.4399 \textcolor{red}{${\downarrow0.01}$} & 0.5881 \textcolor{red}{${\downarrow0.01}$}  & 28.2804 \textcolor{green!70!black}{${\uparrow3.19}$} & 0.8604 \textcolor{green!70!black}{${\uparrow0.04}$} & 0.0896 \textcolor{green!70!black}{${\uparrow0.03}$}  \\
\cdashline{2-10}[1pt/1pt]      & MiniMax~\cite{zi2025minimax} & NeurIPS25& 0.9765  & 0.5889  & 0.4404  & 0.5945  & 29.3671  & 0.8723  & 0.0836 \\
      & YOSE (Minimax) & Ours & 0.9758 \textcolor{blue}{${\downarrow <1e^{-2}}$}  & 0.5778 \textcolor{red}{${\downarrow0.01}$}  & 0.4432 \textcolor{blue}{${\uparrow <1e^{-2}}$} & 0.5953 \textcolor{blue}{${\uparrow <1e^{-2}}$} & 29.5855 \textcolor{green!70!black}{${\uparrow0.22}$} & 0.8703 \textcolor{blue}{${\downarrow <1e^{-2}}$}  & 0.0826 \textcolor{blue}{${\uparrow <1e^{-2}}$}\\
\bottomrule
\bottomrule
\end{tabular}%

} 
    \caption{Quantitative comparison of state-of-the-art VOR methods in VBench~\cite{huang2024vbench} evaluation and background metrics. 
    MiniMax denotes MiniMax Remover. 
    Mot.Smo. means Motion Smooth, Dyn.Deg. denotes Dynamic Degree, Aes.Qua. is Aesthetic Quality, and Ima.Qua. is Imaging Quality.
    \textcolor{green!70!black}{${\uparrow}$}: above the baseline;
    \textcolor{red}{${\downarrow}$}: below the baseline;
    \textcolor{blue}{${\uparrow/\downarrow <1e^{-2}}$}: deviation from baseline is less than ${1e^{-2}}$;
    }
  \label{tab:maintab}%
\end{table*}%

\subsection{Fusion Strategy} 
\label{subsec:fusion}

Although YOSE efficiently reconstructs masked regions, slight boundary inconsistencies may occur between the masked and unmasked areas due to the lack of shared contextual statistics.
To mitigate this, we apply a simple yet effective boundary fusion strategy based on local mean–variance alignment.
Specifically, we first dilate the original mask ${mask}$ to obtain an expanded region ${mask_{dilate}}$ where the generated and original tokens overlap.
We define the overlapping region as ${mask_{overlap}}$:
\begin{equation}
    mask_{overlap}=mask_{dilate}-mask.
\end{equation}
Within this overlapping band ${mask_{overlap}}$, we extract the corresponding part ${Pred_{overlap}}$ from ${St_{in}^{out}}$, ${Orig_{overlap}}$ from ${Video_{mask}}$ by the BVI algorithm.
we compute the mean and standard deviation of both the predicted part ${Pred_{overlap}}$ and original unmasked tokens ${Orig_{overlap}}$, denoted as ${(\mu_{pred}, \sigma_{pred})}$ and ${(\mu_{orig}, \sigma_{orig})}$, respectively.
The predicted region is then normalized and rescaled to match the statistical distribution of the surrounding area:
\begin{gather}
St_{in} = \sigma_{orig} \cdot
\frac{(St_{in} - \mu_{pred})}{\sigma_{pred}} + \mu_{orig}.
\label{eq:fusion}
\end{gather}
After backward mapping, ${out}$ is obtained, which will be combined with ${Video_{mask}}$ by a weighted mask ${\mathcal{M}_{fus}}$:
\begin{equation}
\mathcal{M}_{fus} = \frac{mask+mask_{dilate}}{2},
\label{eq:fusion0}
\end{equation}
\begin{equation}
out=Video_{mask}*(1-\mathcal{M}_{fus})+out*\mathcal{M}_{fus}.
\label{eq:fusion1}
\end{equation}
This operation provides a smooth transition along the mask boundary, effectively alleviating visible seams and enhancing the perceptual coherence between the reconstructed and preserved regions.

\subsection{Loss Setting}
To optimize YOSE, we adopt a flow-matching-based training paradigm that encourages the predicted masked-region output to align with the ground-truth diffusion trajectory.
Given the output ${out}$ from DiffSim and the target noise ${Noise\text{-}GT}$ derived from the forward diffusion process, we define the mask-aware flow matching loss as:
\begin{equation}
\mathcal{L}_{mask}^{FM} =
\frac{| mask \odot (out - (Noise\text{-}GT)) |_2^2}{|mask|_1},
\label{eq:loss}
\end{equation}
where ${\odot}$ denotes element-wise multiplication.
This loss ensures that the diffusion prediction is constrained only within the masked region, promoting precise and efficient object removal while maintaining consistency with the learned diffusion trajectory.
Together with the differentiable BVI and context-aware DiffSim, YOSE achieves adaptive computation that scales linearly with the masked ratio, enabling efficient and semantically coherent video object removal.

\section{Experiments}
\label{sec:exp}
\textbf{Dataset.}
We use VPData~\cite{bian2025videopainter} as the training dataset. 
VPData~\cite{bian2025videopainter} is a large video dataset, containing over 390,000 high-quality video clips (about 866.7 hours of footage). 
Given the relatively small number of parameters to learn, we only used about 70,000 samples from the VPData~\cite{bian2025videopainter} to fit our model.
Each clip includes precise segmentation masks, dense video captions, and detailed descriptions of masked regions.
We only use the video part in VPData~\cite{bian2025videopainter} and train with random masks.
Meanwhile, to evaluate the performance of the models, we employ DAVIS~\cite{DBLP:conf/cvpr/PerazziPMGGS16} and YouTube-VOS~\cite{DBLP:journals/corr/abs-1809-03327} as our evaluation datasets.
Note that YouTube-VOS~\cite{DBLP:journals/corr/abs-1809-03327} contains over 500 short clips, from which we selected 90 longer clips as the test set (equal in number to the DAVIS~\cite{DBLP:conf/cvpr/PerazziPMGGS16} dataset).

\noindent \textbf{Training Details.}
To preserve the model's original capabilities while adapted to YOSE, we only set the three parameters (${\mathcal{G}}$, ${\mathcal{S}}$, ${\mathcal{B}_{bias}}$) trainable.
Our training batch is a total of 32 on 8 GPUs, input frame length is 17, and the resolution is ${480 \times 832}$.
We use AdamW optimizer~\cite{loshchilov2017decoupled} with a learning rate of ${5e^{-5}}$ for 2K training steps within 70K data (takes \textbf{just one epoch}, about 4 hours in total).

\noindent \textbf{Metrics.}
We use four key metrics in VBench~\cite{huang2024vbench} for video generation quality evaluation. 
We also use referenced evaluation metrics, PSNR, SSIM, and LPIPS~\cite{DBLP:conf/cvpr/ZhangIESW18}, to test the quality of the background part.
Considering the varying resolutions each model can handle, we standardize the test resolution to ${480 \times 832}$ and the frame number to ${4\mathbb{N}+1}$, where ${\mathbb{N}}$ denotes an integer.

\begin{figure*}[t]
  \centering
   \includegraphics[width=1\linewidth]{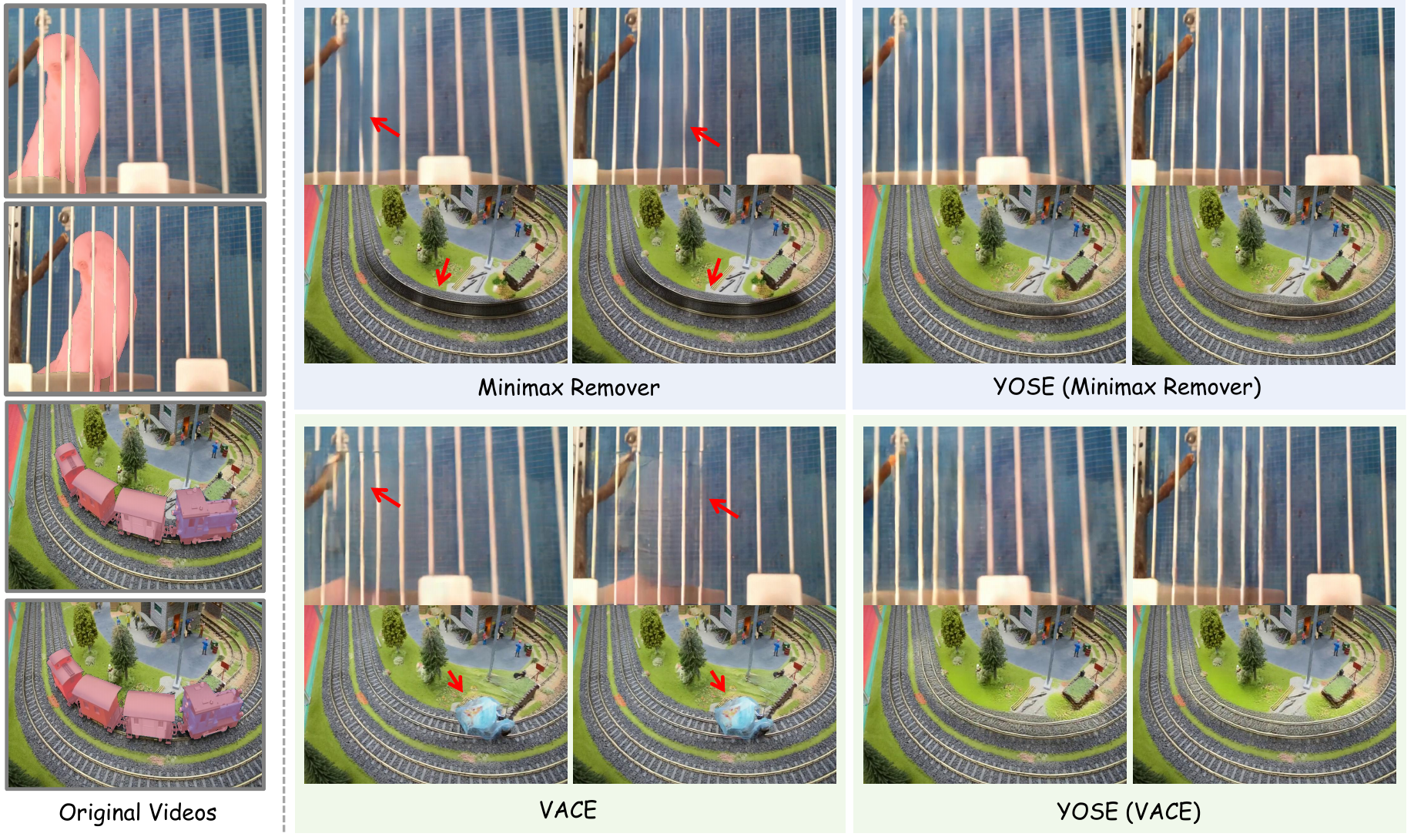}
   \caption{Visual Comparison between YOSE and other methods. The red-masked area is the target to be removed.}
   \label{fig:fig5}
\end{figure*}

\subsection{Mask Ratios in Scenarios}
We have statistically analyzed the distribution of mask ratios within the datasets, DAVIS~\cite{DBLP:conf/cvpr/PerazziPMGGS16} and YouTuBe-VOS~\cite{DBLP:journals/corr/abs-1809-03327}.
It can be observed that the vast majority of samples fall within low mask ratios, with \textbf{around 70\%} of cases having less than 20\% masked area, and only a small fraction exceeding 40\%.
This indicates that real-world video object removal scenarios are typically sparse, where the target object occupies a relatively small portion of the frame.
YOSE is inherently mask-aware, dynamically allocating computation only to masked regions, and achieves increasingly significant acceleration while maintaining reconstruction fidelity, demonstrating clear advantages under realistic usage conditions.

\subsection{FLOPs and Running Time}
Following Minimax remover~\cite{zi2025minimax}'s setting, we define ${[b, n, c]}$ as the token size, ${\eta }$ as the number of DiT, ${h}$ as the number of attention heads, ${f}$ as the dim of FFN layers, and ${\gamma}$ as the mask ratio.
Therefore, the computational complexity FLOPs of YOSE can be expressed as ${\mathbb{G}(\gamma)}$:
\begin{equation}
\mathbb{G}(\gamma) = \gamma \times(49 + 12c + 4n + \frac{{4hn}}{c} + 9f) \beta\eta+....
\label{eq:flops2}
\end{equation}
where ${\beta=b\times n\times c}$.
We also present the visualization curve in Fig.~\ref{fig:fig2}, we set ${b=1}$, ${n=89,040}$, ${c=1,536}$, ${f=8,960}$, ${h=12}$, ${\eta=30}$, for 81-frames 480P video.
${\mathbb{G}(\gamma)}$ exhibits a linear relationship with the mask ratio ${\gamma}$, which demonstrates that YOSE achieves a mask-aware linear computational complexity.

We further evaluate the practical efficiency of the proposed method in terms of runtime performance. 
As shown in Fig.~\ref{fig:fig2} (solid line), the latency curves exhibit a clear mask-dependent trend.
When the mask ratio ${\gamma}$ is 5\%, YOSE achieves a 3.3$\times$ acceleration compared to the full-token DiT(without cross attention) baseline.
At ${\gamma=20\%}$, which corresponds to most real-world video object removal scenarios, our method still provides a 2.5$\times$ speedup.
Moreover, due to the block-encoding constraint of the VAE, when $\gamma$ increases to 80\%, the computation covers all tokens, and the runtime converges to that of full-token inference.
This observation demonstrates that YOSE’s efficiency scales proportionally with the mask ratio, and in the worst case, it does not perform slower than the baseline without YOSE.

\subsection{Quantitative Comparison}
We evaluate the effectiveness and generalizability of YOSE by applying it to two DiT-based video object removal models: MiniMax Remover~\cite{zi2025minimax} and VACE~\cite{jiang2025vace}.
As summarized in Tab.~\ref{tab:maintab}, YOSE consistently preserves or even slightly improves the reconstruction quality while achieving substantially higher computational efficiency.
Notably, in some metrics, YOSE surpasses the base model, benefiting from our design that prevents unnecessary modifications to unmasked regions (also observed in the visual results of Fig.~\ref{fig:fig5}).
This advantage arises from the proposed DiffSim module, which simulates diffusion process information from unmasked regions, allowing the model to maintain unmasked regions.

For MiniMax Remover, YOSE (Minimax) achieves nearly identical perceptual performance to the original full-token model. 
On YouTube-VOS, it maintains comparable Motion Smoothness and Aesthetic Quality while slightly improving background PSNR (from 30.33 to 31.01 dB). 
On DAVIS, the differences across all VBench and background metrics are negligible, confirming that our mask-aware indexing and DiffSim modules introduce only minimal quality loss.
Combined with its observed 2.5 – 3.3 $\times$ inference speedup, YOSE transforms MiniMax Remover into a more efficient and scalable DiT-based VOR model, maintaining the strong generative performance of the original DiT backbone while significantly enhancing runtime practicality.

Furthermore, YOSE demonstrates a distinct advantage in preventing unintended modifications to unmasked background regions, which is particularly evident when applied to VACE.
While the original VACE struggles with background fidelity during generation, YOSE (VACE) boosts background PSNR by an impressive 5.47 dB and SSIM by 0.07 on YouTube-VOS.
Similarly, on the DAVIS dataset, YOSE (VACE) improves PSNR by 3.19 dB.

\begin{table}[t]
  \centering
  \scalebox{0.82}
  {
\begin{tabular}{ccccccc}
\toprule
\multirow{2}[4]{*}{Nis.} & \multirow{2}[4]{*}{Ma.} & \multirow{2}[4]{*}{Fus.} & \multicolumn{2}{c}{VBench~\cite{huang2024vbench} Evaluation} & \multicolumn{2}{c}{Background} \\
\cmidrule{4-7}      &       &       & Dyn.Deg. ${\uparrow}$ & Aes.Qua. ${\uparrow}$ & PSNR ${\uparrow}$ & SSIM ${\uparrow}$ \\
\midrule
$\surd$ & $\surd$ & $\surd$ & \textbf{0.5778 } & \textbf{0.4432 } & \textbf{29.5855 } & \textbf{0.8703 } \\
$\surd$ & $-$   & $\surd$ & 0.5444  & 0.4316  & 27.6164  & 0.8559  \\
$-$   & $\surd$ & $\surd$ & \textbf{0.5778}  & 0.4382  & 27.6686  & 0.8559  \\
$\surd$ & $\surd$ & $-$   & 0.5667  & 0.4320  & 28.4222  & 0.8606  \\
\bottomrule
\bottomrule
\end{tabular}%
} 
    \caption{Ablation Study for DiffSim and Fusion. The best results are \textbf{boldfaced}.
    Nis. means YOSE with ${Lat_{Nis}}$, Ma. denotes YOSE with ${Lat_{mask}}$, and Fus. means with the fusion strategy.}
  \label{tab:abl}%
\end{table}%

\subsection{Qualitative Results}
As shown in Fig.~\ref{fig:fig5}, we present a qualitative comparison between YOSE and the baseline MiniMax Remover on various video object removal cases.
Both methods successfully complete the removal task and maintain global scene consistency.
In particular, the regions reconstructed by YOSE have more coherent texture continuity with the surrounding unmasked areas.
This improvement benefits from the proposed DiffSim module, which simulates diffusion process information from unmasked regions, ensuring external information remains unchanged by DiT processing.
Overall, the visual results suggest that YOSE attains comparable or even superior perceptual quality to MiniMax Remover, while offering much faster and more efficient inference.

\subsection{Ablation Study}
To validate the effectiveness of each component in YOSE, we conduct ablation studies by integrating our method into Minimax Remover~\cite{zi2025minimax} and evaluating on the DAVIS dataset.
We systematically analyze the contributions of the BVI, the DiffSim module, and the Fusion strategy.

\noindent \textbf{BVI Algorithm.} 
As a differentiable token operator, BVI is specifically designed to support multi-batch training under variable-length masks, which greatly improves training efficiency.
With BVI, YOSE can be trained using a batch size of 4 on 8 GPUs in approximately 4 hours.
In contrast, conventional single-batch approaches that cannot handle variable-length tokens require around 11 hours under the same setting, yielding nearly a \textbf{3$\times$ acceleration} in training efficiency for the same batch size.

\begin{figure}[t]
  \centering
   \includegraphics[width=1\linewidth]{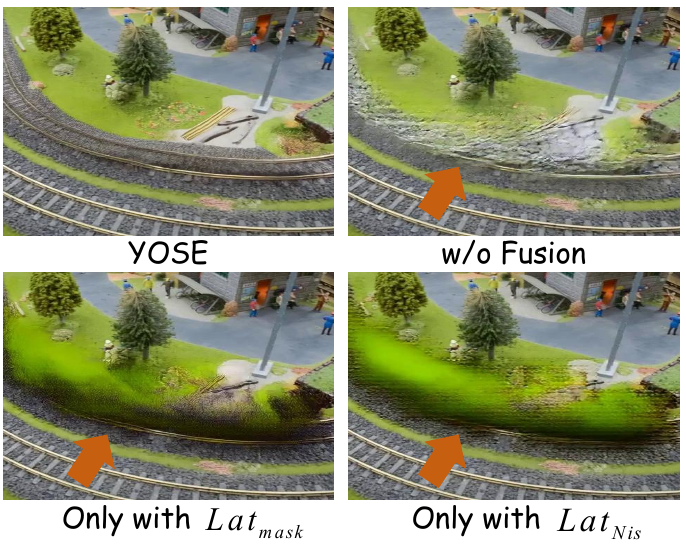}
   \caption{Visual Comparison of Ablation Study for the DiffSim module and the fusion strategy.}
   \label{fig:fig7}
\end{figure}

\noindent \textbf{DiffSim Module.}
As presented in Tab.~\ref{tab:abl} and Fig.~\ref{fig:fig7}, both ${Lat_{Nis}}$ and ${Lat_{mask}}$ contribute to the core diffusion process modeling element of the DiffSim module (detail in Eq.~\eqref{eq:kv}). 
Using only one of them (\ie, Nis. or Ma.) results in a clear drop in both VBench and background reconstruction metrics, indicating insufficient diffusion process simulation.
By jointly leveraging ${Lat_{Nis}}$ and ${Lat_{mask}}$ through the learnable parameters, YOSE achieves the best overall performance, demonstrating the effectiveness of the Diffusion Process Simulator.
 
\noindent \textbf{Fusion Strategy.}
As shown in Tab.~\ref{tab:abl}, removing the proposed fusion strategy leads to slightly lower scores and visible boundary artifacts in Fig.~\ref{fig:fig7}. 
The improvement brought by the fusion strategy confirms its role in harmonizing the transition between restored and preserved regions, effectively reducing edge discontinuities and enhancing perceptual smoothness.

\section{Conclusion}
In this paper, we introduced YOSE, an efficient fine-tuning framework for DiT-based video object removal that leverages mask-aware computation to accelerate both training and inference.
By integrating BVI for differentiable token selection and the DiffSim module for diffusion process contextual consistency, YOSE achieves linear computational complexity with respect to the masked area while maintaining visual quality comparable to full-token models.
\label{sec:conc}

{
{
\clearpage
\setcounter{page}{1}
\maketitlesupplementary

\section{Overview}
This supplementary material provides additional technical details and algorithmic descriptions that complement the main paper.

\section{Mask Embedding}

Diffusion-based video editing frameworks typically rely on a 3D-VAE~\cite{wan2025wan} to encode videos into spatiotemporal latent tokens.
However, existing approaches such as VACE~\cite{jiang2025vace} directly resize the input mask to the latent resolution, which implicitly assumes a pixel-wise correspondence between the RGB space and the latent space.
Since the 3D-VAE encodes the video in spatiotemporal blocks with strides $(F_v, H_v, W_v)$, a simple resize operation does not correctly reflect whether a latent token corresponds to a masked region.
As a result, the latent-space mask generated through resizing may incorrectly mark clean regions as masked or miss fine-grained masked areas.

To obtain an accurate latent-space mask that is consistent with the 3D-VAE’s block-wise encoding, we introduce a block-wise mask embedding mechanism.
Instead of resizing, we determine whether each latent token should be marked as masked by examining the corresponding entire spatiotemporal block in the input mask.
Specifically, for every VAE block of size $(F_v, H_v, W_v)$, we aggregate the binary mask values within the block using a multiplicative rule:
\begin{equation}
1-\prod\limits_{i = 0}^{F_vH_vW_v} {(1 - m_{i}} ),
\label{eq:fusion0}
\end{equation}
where $m_i$ is the binary mask at the $i$-th position inside the block.
This rule ensures that a latent token is considered masked if and only if any pixel inside its receptive block is masked.

The resulting latent-space mask is fully aligned with the 3D-VAE’s tokenization scheme and can seamlessly integrate with our BVI algorithm.
Together, they enable precise identification of essential tokens for selective processing in YOSE.
The full procedure is summarized in Algorithm~\ref{alg:maskemb}.

\begin{algorithm}[h]
    \caption{Mask Embedding}
    \label{alg:maskemb}
    \renewcommand{\algorithmicrequire}{\textbf{Input:}}
    \renewcommand{\algorithmicensure}{\textbf{Output:}}
    
    \begin{algorithmic}[1]
        \REQUIRE ${mask \in \mathbb{R}^{B \times 1 \times F \times H \times W}}$
        \ENSURE ${Emb_{mask}}$
        \STATE ${F_v, H_v,W_v}$: 3D-VAE Stride
        \STATE ${Num=F_v \times H_v \times W_v}$
        \STATE ${mask\mathop  \Rightarrow \limits^{reshape} }$
        \STATE ${\qquad Emb_{mask}\in \mathbb{R}^{B \times 1 \times \frac{F}{F_v} \times F_v \times \frac{H}{H_v} \times H_v \times \frac{W}{W_v} \times W_v }}$
        \STATE ${Emb_{mask} \mathop  \Rightarrow \limits^{transpose} }$
        \STATE ${\qquad Emb_{mask}\in \mathbb{R}^{B \times 1 \times \frac{F}{F_v} \times \frac{H}{H_v}  \times \frac{W}{W_v}  \times F_v \times H_v \times W_v }}$
        \STATE ${Emb_{mask} \mathop  \Rightarrow \limits^{reshape} }$
        \STATE ${\qquad Emb_{mask}\in \mathbb{R}^{B \times 1 \times \frac{F}{F_v} \times \frac{H}{H_v}  \times \frac{W}{W_v}  \times Num }}$
        \STATE ${Emb_{mask} = \prod\limits_{i = 0}^{Num - 1} {(1 - Em{b_{mask}}[...,i]} )}$
        \RETURN ${1-Emb_{mask}}$
    \end{algorithmic}
\end{algorithm}

\section{Details of DiffSim Module}
Algorithm~\ref{alg:ProcSimu} provides the detailed pseudocode of our Diffusion Process Simulator (DiffSim).
DiffSim is designed to simulate the diffusion process in DiT while avoiding a full forward pass through all tokens.
Given the noise latent ${Lat_{Nis}}$, the masked-video latent${Lat_{mask}}$, and the corresponding mask ${mask}$, DiffSim produces the selective latent update used by YOSE.

\noindent \textbf{Mask–guided token partitioning. }
We first convert the input mask into a latent-space mask using the Mask Embedding function (Alg.~\ref{alg:maskemb}).
The latent tokens are then divided into foreground (to be updated) and background (to be preserved) subsets using the Batch Variable-length Indexing function (${BIndex(\cdot)}$, detailed in Alg.~\ref{alg:batchindex}).
This step yields four index sets. ${Ind_{F\_in}, Ind_{F\_out}}$: indices for tokens fed into the short branch, ${Ind_{B\_in}, Ind_{B\_out}}$: indices for tokens sampled back into the full-resolution latent.
This partitioning ensures that only tokens relevant to the edited region will be handled by the DiT-like module.

As shown in Eq.~\eqref{eq:maineq}, we compute the residual latent ${Res_{Nis}}$.
Then, using ${GSample(\cdot)}$, we extract the tokens corresponding to the foreground region, $St_{in}$. 

At each iteration, the current latent features are combined with positional information and passed through a DiT block consisting of attention and feed-forward layers. 
The attention mechanism incorporates information from the simulated latent contexts, while learnable scaling and bias parameters modulate the intermediate representations to better align them with the statistics of the target domain (${St_{in}}$).
Throughout this process, the model repeatedly evaluates which tokens require additional updates based on the mask embedding and the indexing strategy used earlier. Only those latent tokens that correspond to the masked regions are regenerated and refined. Unmasked tokens remain fixed, ensuring stability and avoiding unnecessary computation.
The refinement proceeds iteratively until all DiT blocks have been applied. As a result, the output latent becomes increasingly consistent with both the valid video content and the expected diffusion trajectory, enabling the final reconstruction to better match the underlying spatial-temporal structures of the masked video.

\begin{algorithm}[h]
    \caption{Diffusion Process Simulator}
    \label{alg:ProcSimu}
    \renewcommand{\algorithmicrequire}{\textbf{Input:}}
    \renewcommand{\algorithmicensure}{\textbf{Output:}}
    
    \begin{algorithmic}[1]
        \REQUIRE ${Lat_{Noise}}$,${Lat_{mask}}$, ${mask}$
        \ENSURE ${out}$
        \STATE ${Num_D}$: The Number of DiT Blocks
        \STATE ${\mathcal{G}}$: Learnable Combining Parameters
        \STATE ${\mathcal{S}}$: Learnable Scaling Parameters
        \STATE ${\mathcal{B}_{ias}}$: Learnable Bias Parameters
        \STATE ${Lat_{Nis}}$: The Input Noise Latent
        \STATE ${Lat_{mask}}$: The Latent of Masked Video
        \STATE ${Pos_{emb}}$: Position Embedding
        \STATE ${BIndex()}$: Function of Batch Varible-length Indexing (As Mentioned in Alg.~\ref{alg:batchindex})
        \STATE ${GSample()}$: Function `F.grid\_sample' in Torch 
        \STATE ${Mask\_Emb()}$: Function of Mask Embedding 
        \STATE ${mask=Mask\_Emb(mask)}$
        \STATE ${Ind_{F\_in}, Ind_{B\_in} = BIndex(mask)}$
        \STATE ${Ind_{F\_out}, Ind_{B\_out} = BIndex(1-mask)}$
        \STATE ${Res_{Nis}=Lat_{Nis}-Lat_{mask}}$
        \STATE ${St_{in} = GSample(Lat_{mask}, Ind_{F\_in})}$
        \FOR{each $i \in [0, Num_D-1]$}
            \STATE ${Q=Apply(St_{in}, GSample(Pos_{emb}, Ind_{F\_in}))}$ 
            \STATE ${KV=\mathcal{G}[i]*Lat_{mask}+(1-\mathcal{G}[i])*Res_{Nis}}$
            \STATE ${KV = (1+\mathcal{S}[i])*KV+\mathcal{B}_{ias}[i]}$
            \STATE ${KV = Apply(KV\cup St_{in}, Pos_{emb})}$
            \STATE ${St_{in}=Attn\&FFN(Q, K, V)}$
        \ENDFOR
        \STATE ${out = mask*GSample(St_{in}, Ind_{B\_in})}$
        \RETURN ${out}$
    \end{algorithmic}
\end{algorithm}

\section{More Details}

\subsection{Causal Encoding of 3D-VAE}
Previous study~\cite{wu2025improved} found that the slight localized blurring and color casts in certain cases were primarily artifacts of the 3D-VAE’s causal encoding mechanism. 
This mechanism can introduce statistical inconsistencies at mask boundaries during the latent space transformation, particularly when the content inside the mask is not aligned with its surroundings. 
To resolve this, YOSE pre-fills masked regions with neighboring pixels prior to encoding, thereby harmonizing feature distributions across boundaries. 

\begin{figure}[h]
  \centering
   \includegraphics[width=1\linewidth]{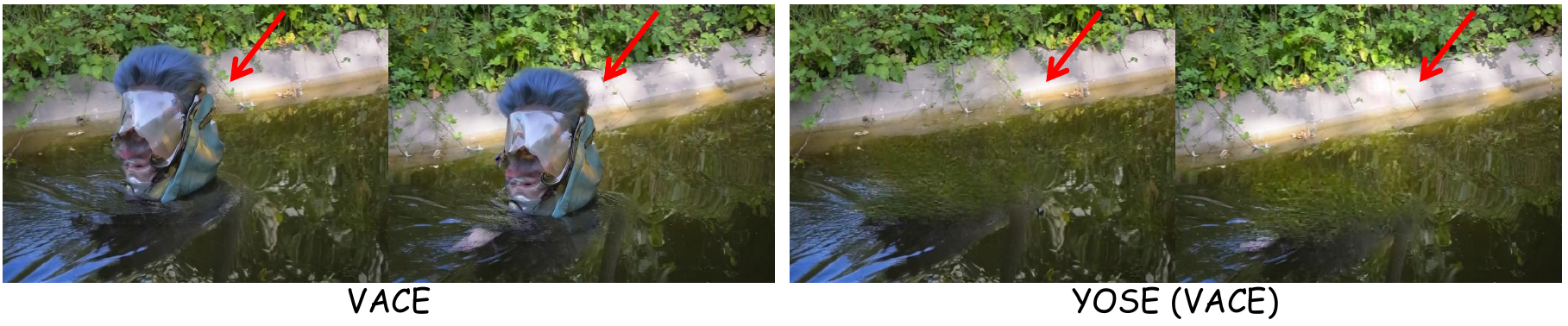}
   \vspace{-0.6cm}
   \caption{Visual Comparison between VACE and YOSE (VACE).}
   \vspace{-0.6cm}
   \label{fig:sup2}
\end{figure}

\subsection{Efficiency Metrics}
The Tab.~\ref{tab:fig5} shows a comparison of the speeds of various DiT-based methods.
We applied YOSE to VACE, a ControlNet-like DiT-based video editing model. 
As shown in Fig.~\ref{fig:sup2}, the original VACE suffers from mask-shaped semantic bias, inadvertently generating mask-shaped foreground objects, and significantly altering the background.
After applying YOSE, we discovered that YOSE’s selective token processing effectively suppresses these unwanted hallucinations.
By focusing computation exclusively on essential tokens within the mask, YOSE prevents the model from over-interpreting mask semantics, thus achieving the improvement of success rate in object removal(from 62.2\% to 97.8\%). 
Since the control branch of VACE consumes significant computation, which cannot be accelerated by YOSE, the acceleration effect of YOSE on this part is relatively limited.

\begin{table}[h]
  \centering
  \scalebox{0.7}[0.72]
  {
\begin{tabular}{cccccc}
\toprule
VideoPainter & ROSE  & VACE & YOSE (VACE) & Minimax & YOSE (Minimax) \\
\midrule
 0.402  & 1.066  & 0.308 & 0.417 & 9.515  & 24.509  \\
\bottomrule
\end{tabular}%
}
\vspace{-0.2cm}
 \caption{Efficiency Metrics (FPS).}
 \vspace{-0.5cm}
  \label{tab:fig5}%
\end{table}

\subsection{Relative Performance Change Analysis}
As shown in the scatter plot (Fig.~\ref{fig:sup3}), we measured the Relative Performance Change (${\gamma}$ /\%) across 180 cases in two datasets. 
$76.1\%$ of cases fall within the mask ratio range of 0–25\%, exhibiting minimal fluctuations in visual quality.
As the mask ratio increases, the amplitude of fluctuations begins to grow, ${\pm 25\%}$.
Notably, the vast majority of cases, about 75\%, fluctuate within a range of ${\pm 5\%}$, which demonstrates YOSE's high robustness and stability across diverse removal scenarios.

\begin{figure}[h]
  \centering
   \includegraphics[width=1\linewidth]{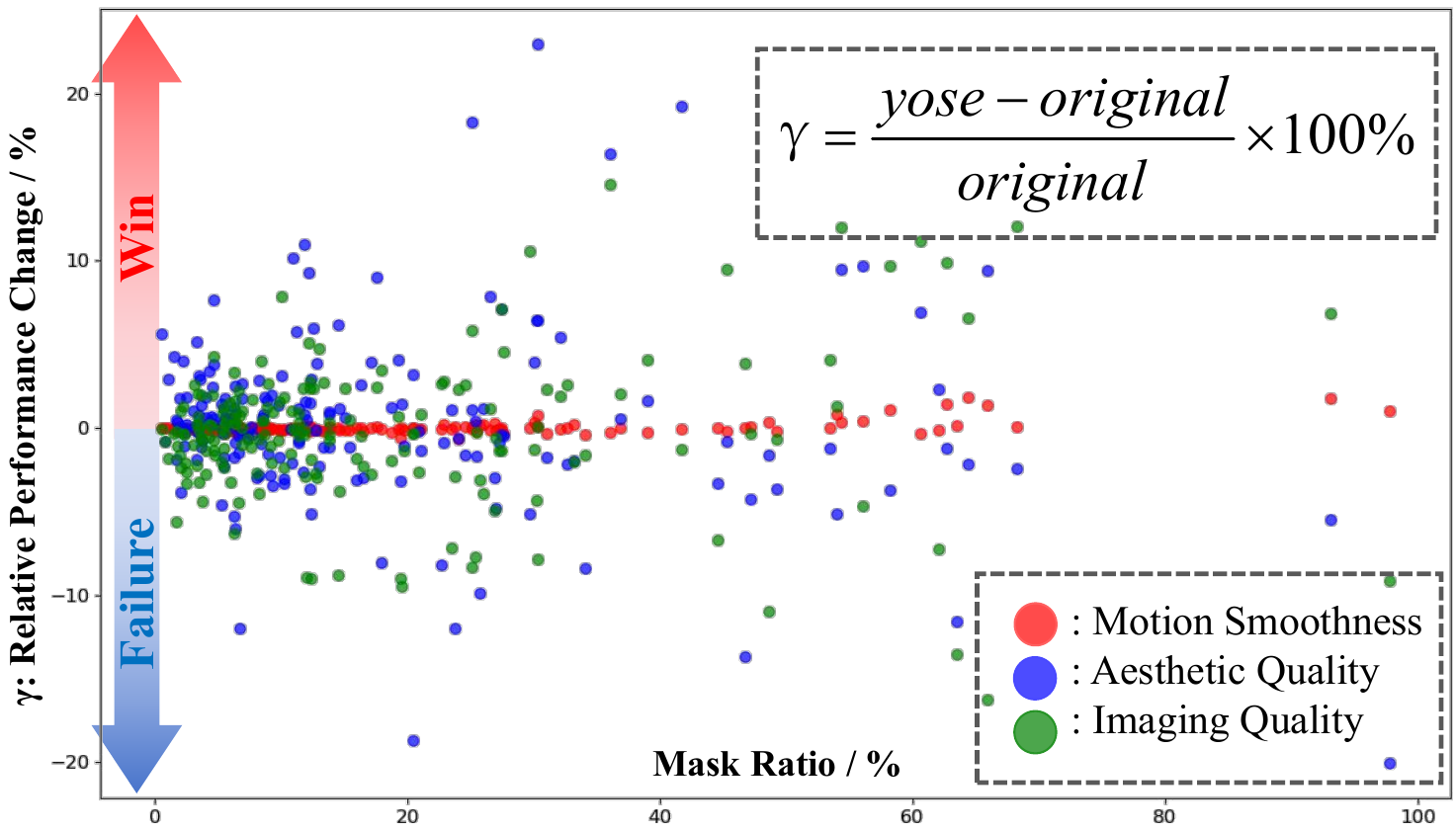}
   \vspace{-0.5cm}
   \caption{Relative Performance Change Analysis.}
   \vspace{-0.5cm}
   \label{fig:sup3}
\end{figure}
}
}

{
    \small
    \bibliographystyle{ieeenat_fullname}
    \bibliography{main}

\begin{thebibliography}{28}
\providecommand{\natexlab}[1]{#1}
\providecommand{\url}[1]{\texttt{#1}}
\expandafter\ifx\csname urlstyle\endcsname\relax
  \providecommand{\doi}[1]{doi: #1}\else
  \providecommand{\doi}{doi: \begingroup \urlstyle{rm}\Url}\fi

\bibitem[Bian et~al.(2025)Bian, Zhang, Ju, Cao, Xie, Shan, and Xu]{bian2025videopainter}
Yuxuan Bian, Zhaoyang Zhang, Xuan Ju, Mingdeng Cao, Liangbin Xie, Ying Shan, and Qiang Xu.
\newblock Videopainter: Any-length video inpainting and editing with plug-and-play context control.
\newblock In \emph{SIGGRAPH}, pages 1--12, 2025.

\bibitem[Duan et~al.(2025)Duan, Zhang, Jin, Zhang, Xiong, Zou, Ren, Guo, and Li]{duan2025dit4sr}
Zheng-Peng Duan, Jiawei Zhang, Xin Jin, Ziheng Zhang, Zheng Xiong, Dongqing Zou, Jimmy~S Ren, Chunle Guo, and Chongyi Li.
\newblock Dit4sr: Taming diffusion transformer for real-world image super-resolution.
\newblock In \emph{ICCV}, pages 18948--18958, 2025.

\bibitem[Esser et~al.(2024)Esser, Kulal, Blattmann, Entezari, M{\"u}ller, Saini, Levi, Lorenz, Sauer, Boesel, et~al.]{esser2024scaling}
Patrick Esser, Sumith Kulal, Andreas Blattmann, Rahim Entezari, Jonas M{\"u}ller, Harry Saini, Yam Levi, Dominik Lorenz, Axel Sauer, Frederic Boesel, et~al.
\newblock Scaling rectified flow transformers for high-resolution image synthesis.
\newblock In \emph{Forty-first international conference on machine learning}, 2024.

\bibitem[Fang et~al.(2025)Fang, Chen, and Guo]{fengvideodit}
Pengcheng Fang, Yuxia Chen, and Rui Guo.
\newblock When and what: Diffusion-grounded videollm with entity aware segmentation for long video understanding.
\newblock \emph{CoRR}, abs/2508.15641, 2025.

\bibitem[Feng et~al.(2025)Feng, Ma, Wang, Qi, Chen, Chen, and Wang]{feng2025dit4edit}
Kunyu Feng, Yue Ma, Bingyuan Wang, Chenyang Qi, Haozhe Chen, Qifeng Chen, and Zeyu Wang.
\newblock Dit4edit: Diffusion transformer for image editing.
\newblock In \emph{AAAI}, pages 2969--2977, 2025.

\bibitem[Huang et~al.(2024)Huang, He, Yu, Zhang, Si, Jiang, Zhang, Wu, Jin, Chanpaisit, et~al.]{huang2024vbench}
Ziqi Huang, Yinan He, Jiashuo Yu, Fan Zhang, Chenyang Si, Yuming Jiang, Yuanhan Zhang, Tianxing Wu, Qingyang Jin, Nattapol Chanpaisit, et~al.
\newblock Vbench: Comprehensive benchmark suite for video generative models.
\newblock In \emph{CVPR}, pages 21807--21818, 2024.

\bibitem[Jiang et~al.(2025)Jiang, Han, Mao, Zhang, Pan, and Liu]{jiang2025vace}
Zeyinzi Jiang, Zhen Han, Chaojie Mao, Jingfeng Zhang, Yulin Pan, and Yu Liu.
\newblock Vace: All-in-one video creation and editing.
\newblock \emph{ICCV}, 2025.

\bibitem[Kong et~al.(2025)Kong, Li, Wang, Xu, Pei, Li, and Ren]{kong2025dual}
Dehong Kong, Fan Li, Zhixin Wang, Jiaqi Xu, Renjing Pei, Wenbo Li, and WenQi Ren.
\newblock Dual prompting image restoration with diffusion transformers.
\newblock In \emph{CVPR}, pages 12809--12819, 2025.

\bibitem[Li et~al.(2024)Li, Zhang, Huang, Liu, Pei, Shao, and Xu]{li2024magiceraser}
Fan Li, Zixiao Zhang, Yi Huang, Jianzhuang Liu, Renjing Pei, Bin Shao, and Songcen Xu.
\newblock Magiceraser: Erasing any objects via semantics-aware control.
\newblock In \emph{ECCV}, pages 215--231. Springer, 2024.

\bibitem[Li et~al.(2025)Li, Xue, Ren, and Bo]{li2025diffueraser}
Xiaowen Li, Haolan Xue, Peiran Ren, and Liefeng Bo.
\newblock Diffueraser: A diffusion model for video inpainting.
\newblock \emph{arXiv preprint arXiv:2501.10018}, 2025.

\bibitem[Loshchilov and Hutter(2017)]{loshchilov2017decoupled}
Ilya Loshchilov and Frank Hutter.
\newblock Decoupled weight decay regularization.
\newblock \emph{arXiv preprint arXiv:1711.05101}, 2017.

\bibitem[Lu et~al.(2025)Lu, Jiang, Jin, Yang, Gong, Shi, Wang, and Zhao]{lu2025grig}
Wanglong Lu, Xianta Jiang, Xiaogang Jin, Yong-Liang Yang, Minglun Gong, Kaijie Shi, Tao Wang, and Hanli Zhao.
\newblock Grig: Data-efficient generative residual image inpainting.
\newblock \emph{Computational Visual Media}, 11\penalty0 (6):\penalty0 1329--1361, 2025.

\bibitem[Meng et~al.(2025)Meng, Jin, Lei, Guo, and Li]{meng2025ultraled}
Yuang Meng, Xin Jin, Lina Lei, Chun-Le Guo, and Chongyi Li.
\newblock Ultraled: Learning to see everything in ultra-high dynamic range scenes.
\newblock \emph{arXiv preprint arXiv:2510.07741}, 2025.

\bibitem[Miao et~al.(2025)Miao, Feng, Zeng, Gao, Liu, Yan, Qi, Chen, Wang, and Zhao]{miao2025rose}
Chenxuan Miao, Yutong Feng, Jianshu Zeng, Zixiang Gao, Hantang Liu, Yunfeng Yan, Donglian Qi, Xi Chen, Bin Wang, and Hengshuang Zhao.
\newblock Rose: Remove objects with side effects in videos.
\newblock \emph{NeurIPS}, 2025.

\bibitem[Perazzi et~al.(2016)Perazzi, Pont{-}Tuset, McWilliams, Gool, Gross, and Sorkine{-}Hornung]{DBLP:conf/cvpr/PerazziPMGGS16}
Federico Perazzi, Jordi Pont{-}Tuset, Brian McWilliams, Luc~Van Gool, Markus~H. Gross, and Alexander Sorkine{-}Hornung.
\newblock A benchmark dataset and evaluation methodology for video object segmentation.
\newblock In \emph{CVPR}, pages 724--732. {IEEE} Computer Society, 2016.

\bibitem[Qin et~al.(2025{\natexlab{a}})Qin, Wang, Xu, Li, Zhang, and Wan]{qin2025unified}
Mengjie Qin, Wen Wang, Honghui Xu, Te Li, Chunlong Zhang, and Minhong Wan.
\newblock Unified transformed t-svd using unfolding tensors for visual inpainting.
\newblock \emph{Computational Visual Media}, 2025{\natexlab{a}}.

\bibitem[Qin et~al.(2025{\natexlab{b}})Qin, Quan, Chen, and Ji]{qin2025robust}
Xinran Qin, Yuhui Quan, Zhuojie Chen, and Hui Ji.
\newblock Robust unsupervised deep learning for nonblind image deconvolution with inaccurate kernels.
\newblock \emph{TNNLS}, 2025{\natexlab{b}}.

\bibitem[Qin et~al.(2025{\natexlab{c}})Qin, Wang, Li, Chen, Pei, Li, and Cao]{qin2025camedit}
Xinran Qin, Zhixin Wang, Fan Li, Haoyu Chen, Renjing Pei, Wenbo Li, and Xiaochun Cao.
\newblock Camedit: Continuous camera parameter control for photorealistic image editing.
\newblock In \emph{NeurIPS}, 2025{\natexlab{c}}.

\bibitem[Qin et~al.(2026)Qin, Cui, Sun, Chen, Ren, Knoll, and Cao]{qin2026disentangle}
Xinran Qin, Yuning Cui, Shangquan Sun, Ruoyu Chen, Wenqi Ren, Alois Knoll, and Xiaochun Cao.
\newblock Disentangle to fuse: Towards content preservation and cross-modality consistency for multi-modality image fusion.
\newblock \emph{TIP}, 2026.

\bibitem[Wan et~al.(2025)Wan, Wang, Ai, Wen, Mao, Xie, Chen, Yu, Zhao, Yang, et~al.]{wan2025wan}
Team Wan, Ang Wang, Baole Ai, Bin Wen, Chaojie Mao, Chen-Wei Xie, Di Chen, Feiwu Yu, Haiming Zhao, Jianxiao Yang, et~al.
\newblock Wan: Open and advanced large-scale video generative models.
\newblock \emph{arXiv preprint arXiv:2503.20314}, 2025.

\bibitem[Wu et~al.(2025{\natexlab{a}})Wu, Fu, Guo, Han, and Li]{wu2025vtinker}
Chenyang Wu, Jiayi Fu, Chun-Le Guo, Shuhao Han, and Chongyi Li.
\newblock Vtinker: Guided flow upsampling and texture mapping for high-resolution video frame interpolation.
\newblock \emph{arXiv preprint arXiv:2511.16124}, 2025{\natexlab{a}}.

\bibitem[Wu et~al.(2025{\natexlab{b}})Wu, Zhu, Liu, Zhao, Zhai, Cao, and Zha]{wu2025improved}
Pingyu Wu, Kai Zhu, Yu Liu, Liming Zhao, Wei Zhai, Yang Cao, and Zheng-Jun Zha.
\newblock Improved video vae for latent video diffusion model.
\newblock In \emph{CVPR}, pages 18124--18133, 2025{\natexlab{b}}.

\bibitem[Xu et~al.(2018)Xu, Yang, Fan, Yue, Liang, Yang, and Huang]{DBLP:journals/corr/abs-1809-03327}
Ning Xu, Linjie Yang, Yuchen Fan, Dingcheng Yue, Yuchen Liang, Jianchao Yang, and Thomas~S. Huang.
\newblock Youtube-vos: {A} large-scale video object segmentation benchmark.
\newblock \emph{CoRR}, abs/1809.03327, 2018.

\bibitem[Yang et~al.(2024)Yang, Teng, Zheng, Ding, Huang, Xu, Yang, Hong, Zhang, Feng, et~al.]{yang2024cogvideox}
Zhuoyi Yang, Jiayan Teng, Wendi Zheng, Ming Ding, Shiyu Huang, Jiazheng Xu, Yuanming Yang, Wenyi Hong, Xiaohan Zhang, Guanyu Feng, et~al.
\newblock Cogvideox: Text-to-video diffusion models with an expert transformer.
\newblock \emph{arXiv preprint arXiv:2408.06072}, 2024.

\bibitem[Zhang et~al.(2018)Zhang, Isola, Efros, Shechtman, and Wang]{DBLP:conf/cvpr/ZhangIESW18}
Richard Zhang, Phillip Isola, Alexei~A. Efros, Eli Shechtman, and Oliver Wang.
\newblock The unreasonable effectiveness of deep features as a perceptual metric.
\newblock In \emph{CVPR}, pages 586--595. Computer Vision Foundation / {IEEE} Computer Society, 2018.

\bibitem[Zhong et~al.(2025)Zhong, Li, Huang, Liu, Pei, and Song]{zhong2025outdreamer}
Linhao Zhong, Fan Li, Yi Huang, Jianzhuang Liu, Renjing Pei, and Fenglong Song.
\newblock Outdreamer: Video outpainting with a diffusion transformer.
\newblock \emph{arXiv preprint arXiv:2506.22298}, 2025.

\bibitem[Zhou et~al.(2023)Zhou, Li, Chan, and Loy]{zhou2023propainter}
Shangchen Zhou, Chongyi Li, Kelvin~CK Chan, and Chen~Change Loy.
\newblock Propainter: Improving propagation and transformer for video inpainting.
\newblock In \emph{ICCV}, pages 10477--10486, 2023.

\bibitem[Zi et~al.(2025)Zi, Peng, Qi, Wang, Zhao, Xiao, and Wong]{zi2025minimax}
Bojia Zi, Weixuan Peng, Xianbiao Qi, Jianan Wang, Shihao Zhao, Rong Xiao, and Kam-Fai Wong.
\newblock Minimax-remover: Taming bad noise helps video object removal.
\newblock \emph{NeurIPS}, 2025.

\end{thebibliography}
}

\end{document}